\documentclass[11pt]{article}

\usepackage{acl}

\usepackage{times}
\usepackage{latexsym}

\usepackage[T1]{fontenc}
\usepackage[utf8]{inputenc}

\usepackage{microtype}
\usepackage{algorithm}
\usepackage{algpseudocode}
\usepackage{inconsolata}

\usepackage{graphicx}

\usepackage{amsmath}   % per align, \text, \nabla ecc.
\usepackage{amssymb}   % per simboli aggiuntivi come \triangleq
\usepackage{amsfonts}  % (opzionale) per migliorare font matematici come \mathcal

\usepackage{booktabs}

\usepackage{float}
\usepackage{stfloats}

\usepackage{listings}
\usepackage{xcolor}

\usepackage{subcaption}
\usepackage{enumitem}

\lstdefinestyle{promptstyle}{
    basicstyle=\footnotesize\ttfamily,
    frame=single,
    breaklines=true,
    breakatwhitespace=true,
    columns=fullflexible,
    aboveskip=2pt,
    belowskip=2pt,
    framesep=3pt,
    xleftmargin=0pt,
    xrightmargin=0pt,
}

\title{\textsc{GradRAG}: Cross-Component Prompt Adaptation\\
for Coordinated Multi-Agent RAG}

\author{Paolo Pedinotti\thanks{Paolo Pedinotti contributed to this work during his internship at Bloomberg.}\\
  Bloomberg\\
  \texttt{pedinotti.paolo@gmail.com} \\
  \\\And
  Enrico Santus\thanks{Equal contribution.} \\
  Bloomberg \\
  \texttt{esantus@bloomberg.net} \\}

\begin{document}
\maketitle

\begin{abstract}
Retrieval-Augmented Generation (RAG) systems increasingly employ multiple LLM agents. Yet, most prior work optimizes components in isolation rather than coordinating improvements across the pipeline. We introduce \textbf{\textsc{GradRAG}}, a framework for \textbf{cross-component prompt adaptation} that models the RAG pipeline as a computational graph and propagates structured evaluation feedback to update upstream agents. An \emph{Evaluator} critiques downstream answers and supporting evidence, producing actionable feedback that a \emph{Prompt Optimizer} uses to iteratively update adaptive agents (e.g., retrievers, graph constructors, answerers). The Evaluator also triggers early stopping when the output is deemed satisfactory.
We evaluate \textsc{GradRAG} on the \textsc{SQuALITY} and \textsc{QMSum} benchmarks under two retrieval paradigms: 
(i) flat (chunk-based) retrieval using IRCoT-style query refinement \citep{trivedi-etal-2023-interleaving}, and 
(ii) graph-based retrieval that constructs and iteratively enriches an entity–relation graph from the document. 
Across both settings, \textsc{GradRAG} consistently outperforms one-step refinement baselines that update only the final generator, achieving a \textbf{12--15 percentage point net preference margin} in LLM-judged pairwise comparisons, with most gains realized within two refinement iterations.
\end{abstract}

\begin{figure*}[t]
    \centering
    \includegraphics[width=.9\textwidth]{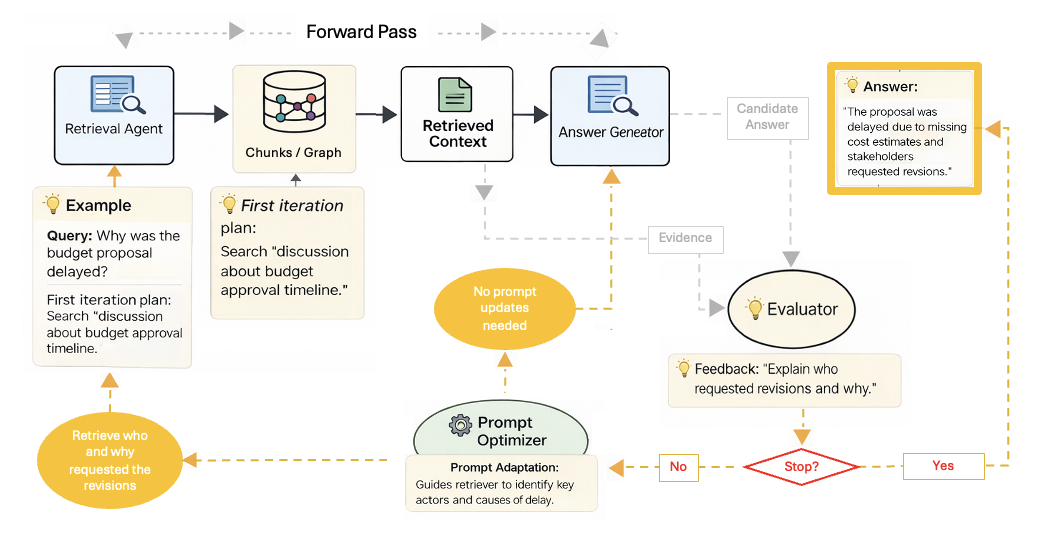}
    %\caption{\textsc{GradRAG}: cross-component prompt adaptation in an agentic RAG pipeline.
    %\textbf{Top:} Document-grounded forward pass for a mock query (``Why was the budget proposal delayed?''), where a retrieval agent plans focused searches and selects evidence from either document chunks or structured graph representations, which are then used by an answer-generation agent.
    %\textbf{Bottom:} An Evaluator reviews the answer and supporting evidence, and a Prompt Optimizer updates prompts across retrieval and generation components, enabling coordinated refinement of the pipeline.}
    \caption{\textsc{GradRAG}: cross-component prompt adaptation in an agentic RAG pipeline.
    \textbf{Grey (forward pass):} A Retrieval Agent gathers evidence from either document chunks (Vector RAG) or a structured entity--relation graph (GraphRAG, constructed by a separate agent that is not shown in the mock up), which is assembled into a context and passed to an Answer Generation Agent to produce a candidate answer.
    \textbf{Orange (evaluation and control):} An Evaluator reviews the answer and its supporting evidence and returns feedback with a binary \emph{Stop?} decision. If \emph{Yes}, the answer is accepted. If \emph{No}, a Prompt Optimizer updates the prompts of one or more agents (as indicated by the evaluation), and the next forward pass is executed.}

    \label{fig:mockup}
\end{figure*}

\section{Introduction}

Retrieval-Augmented Generation (RAG) has become a central paradigm for enabling large language models (LLMs) to reason over information beyond their parametric knowledge. Early RAG systems followed a simple \emph{retrieve-then-generate} workflow in which a retriever selects relevant text and a generator produces an answer conditioned on it. More recently, RAG architectures have evolved into increasingly \emph{agentic} systems, where multiple LLM-based components cooperate to formulate retrieval queries, organize evidence, perform multi-step reasoning, and evaluate or revise intermediate outputs.

Despite these advances, most RAG pipelines still optimize components in isolation. Refinements are typically applied locally—for example, through query rewriting, evidence filtering, or post-hoc answer editing via \textit{self-reflection} or \textit{self-correction}. Consequently, errors introduced early in the pipeline, such as incomplete retrieval or poorly structured evidence, often propagate downstream and limit overall performance. This reveals a coordination problem in multi-agent RAG pipelines: evaluation signals produced at the final generation stage are rarely used to improve upstream components such as retrievers or graph construction modules.

In this work, we study document-grounded query answering and query-focused summarization, where retrieval operates within a given document (or document set) by selecting relevant spans for each query. This setting corresponds to benchmarks such as \textsc{SQuALITY} and \textsc{QMSum}, where each query is paired with long documents. Although these documents fit within modern LLM context windows, our goal is not to compare RAG with full-context inference but to use these benchmarks as a controlled testbed for analyzing how different RAG pipelines behave and how coordination between their components can be improved.

We introduce \textbf{\textsc{GradRAG}}, a framework for \textbf{cross-component prompt adaptation} in agentic RAG systems (Figure~\ref{fig:mockup}). \textsc{GradRAG} models the RAG pipeline as a computational graph whose nodes correspond to agents such as the retriever, graph constructor, and answer generator. Instead of refining only the final answer, the framework propagates evaluation feedback to multiple upstream agents responsible for retrieval and evidence construction. At test time, an \emph{Evaluator} agent analyzes the generated answer together with its supporting evidence and produces structured feedback identifying missing information, weak reasoning links, or irrelevant context. A \emph{Prompt Optimizer} converts this feedback into prompt updates for adaptive agents across refinement iterations. The process operates entirely through prompt updates and does not modify model parameters.

By allowing downstream critiques to influence upstream decisions, \textsc{GradRAG} enables coordinated improvements beyond one-step answer rewriting. Rather than optimizing retrieval, structuring, and generation independently, the framework aligns their behavior through feedback-driven prompt refinement within a bounded test-time budget.

We evaluate \textsc{GradRAG} on \textsc{SQuALITY} and \textsc{QMSum} using two representative retrieval paradigms. The first is a \textbf{flat (chunk-based) retrieval} pipeline, where the system iteratively rewrites search queries and retrieves document chunks using a hybrid dense--lexical retriever following an IRCoT-style strategy \citep{trivedi-etal-2023-interleaving}. The second is a \textbf{graph-based retrieval} pipeline that extracts and iteratively enriches an entity–relation graph to assemble evidence for answering the query. Within each paradigm we compare GradRAG against controlled one-step refinement baselines in which only the final answer generator is updated, isolating the effect of cross-component prompt adaptation. Across both settings, \textsc{GradRAG} consistently outperforms these baselines, with most gains realized within the first two refinement iterations. Section~\ref{sec:res} shows that cross-component prompt adaptation yields a \textbf{12–15 percentage point net preference margin} in LLM-guided pairwise evaluations. \textsc{GradRAG} therefore provides a general framework for coordinating improvements across heterogeneous components in modular RAG pipelines. This work makes the following contributions:
\begin{itemize}
    \item We introduce a framework for \textbf{cross-component prompt adaptation} in agentic RAG, enabling evaluator feedback to update the prompts of multiple upstream agents at test time.
    \item We show that \textbf{coordinated agent refinement} via shared evaluative feedback improves alignment between retrieval, structuring, and generation.
    \item We propose a \textbf{modular graph-based abstraction} that applies the same feedback-driven optimization across both flat (chunk-based) and graph-based RAG pipelines.
\end{itemize}

\section{Related Work}
\label{sec:Related-work}

\subsection{Agentic Approaches and RAG}

Recent work in Retrieval-Augmented Generation (RAG) increasingly incorporates agentic components. \citet{asai2023selfraglearningretrievegenerate} introduce a \emph{reflection-based mechanism} in which the model produces intermediate tokens representing retrieval decisions and evidence assessments, enabling iterative retrieval and answer refinement guided by a learned critic. This work established that RAG performance can benefit from multi-step, agentic decision-making.

Subsequent approaches distribute retrieval and reasoning across specialized components. For example, \citet{jeong-etal-2024-adaptive} and \citet{lee-etal-2025-hybgrag} propose routing agents that decide whether and how extensively to retrieve and which backend to query. \citet{trivedi-etal-2023-interleaving} frame retrieval as a sequential decision process that iteratively refines search queries, while \citet{jiang-etal-2025-rag} use Monte Carlo Tree Search guided by learned reward models. In this literature, \emph{multi-agent} typically refers to pipelines composed of multiple LLM calls with distinct prompts, each specializing in a subtask such as retrieval routing, evidence assessment, or answer synthesis.

Multi-agent retrieval optimization has become a common strategy. \citet{yan2024correctiveretrievalaugmentedgeneration} train a relevance judge that decides whether to continue querying or switch to generation, while \citet{chang-etal-2025-main} generate partial answers from retrieved documents and use a judge to select the most informative ones. Despite their success, these approaches apply feedback locally, focusing on individual stages such as retrieval control or answer selection. Consequently, evaluation signals generated at the final answer stage rarely influence upstream components such as retrieval or evidence construction. In contrast, our work addresses this coordination problem directly: how feedback on downstream answer quality can be used to adapt multiple upstream components of a RAG pipeline in a unified manner.

\subsection{LLMs as Critic Agents and Prompt Optimizers}

The \emph{reflection} pattern can be viewed as a three-stage loop of generation, evaluation, and adaptation. Early systems \citep{10.5555/3666122.3668141,10.5555/3666122.3666499} demonstrated inference-time refinement in which models iteratively revise outputs using self- or evaluator-generated critiques (later described as \emph{textual gradients} by \citet{yuksekgonul2024textgradautomaticdifferentiationtext}, or \emph{verbal reinforcement}). These approaches treat natural-language critiques as optimization signals guiding subsequent reasoning steps.

This idea was later generalized into \textbf{automatic prompt optimization}, where an LLM searches for prompts that improve downstream performance \cite{zhou2023largelanguagemodelshumanlevel,opsahl-ong-etal-2024-optimizing}. Frameworks such as DSPy \cite{khattab2023dspycompilingdeclarativelanguage} and TextGrad \cite{yuksekgonul2024textgradautomaticdifferentiationtext} provide programming abstractions for this process: DSPy focuses on optimizing prompting structure and demonstrations, while TextGrad formalizes critique-driven prompt updates over computational graphs.

Our work builds on these foundations but focuses specifically on coordinating prompt adaptation across multiple components of a RAG pipeline. \textsc{GradRAG} instantiates critique-driven prompt optimization within agentic RAG systems and empirically evaluates how coordinating updates across retrieval, structuring, and generation components improves retrieval-augmented reasoning over long documents.

\section{Method: \textsc{GradRAG}}
\label{sec:method}

\textbf{\textsc{GradRAG}} is a test-time framework for coordinating multiple agents in Retrieval-Augmented Generation (RAG) pipelines through \textbf{cross-component prompt adaptation}. \textsc{GradRAG} builds on critique-driven prompt optimization and reflection-based approaches, instantiating them in agentic RAG systems where feedback on downstream answer quality is used to adapt multiple upstream components in a unified manner.

\paragraph{Overview.}
As illustrated in Figure~\ref{fig:mockup}, \textsc{GradRAG} executes a RAG pipeline in iterative refinement cycles. Each cycle consists of a forward pass that produces intermediate artifacts (e.g., retrieved text chunks or extracted graph structures) and a final answer, followed by an evaluation step. An \emph{Evaluator} agent reviews the answer together with its supporting evidence and produces structured natural-language feedback. If refinement is required, a \emph{Prompt Optimizer} converts this feedback into updated prompts for one or more adaptive agents. The next cycle then runs with the updated prompts. This process operates entirely through prompt updates and does not modify model parameters. The full refinement loop is summarized in Algorithm~\ref{alg:gradrag}.

\paragraph{Computational graph abstraction.}
\textsc{GradRAG} models the RAG workflow as a computational graph whose nodes represent agents, programs, or data artifacts, and whose edges represent computational transformations. Nodes corresponding to LLM-based agents are associated with prompts, which may be either fixed or adaptive. In our framework, \textbf{adaptive agents} are those whose prompts can be updated across refinement cycles based on evaluator feedback. This abstraction allows heterogeneous components — such as classical retrievers and LLM-based agents — to coexist within the same pipeline.

\paragraph{Cross-component prompt adaptation.}
Unlike one-step reflection approaches that refine only the final answer generator, \textsc{GradRAG} allows evaluator feedback to update the prompts of multiple upstream agents. For example, feedback indicating missing causal or explanatory information may trigger changes to both the retrieval strategy (to target more relevant evidence) and the answer generation style (to emphasize causal structure). Prompt updates are applied iteratively within a bounded test-time budget, enabling coordinated improvements across components.

\paragraph{Early stopping.}
Following prior reflection-based architectures \citep{10.5555/3666122.3668141,10.5555/3666122.3666499}, \textsc{GradRAG} employs an early-stopping mechanism. After each evaluation step, the Evaluator returns a discrete decision indicating whether the current output is satisfactory. Refinement cycles stop when the Evaluator signals satisfaction or when a maximum number of iterations is reached (Algorithm~\ref{alg:gradrag}, lines~6--8).

\begin{algorithm}[t]
\small
\caption{\textsc{GradRAG}: Cross-Component Prompt Adaptation}
\label{alg:gradrag}
\begin{algorithmic}[1]
\Require Query $q$, document(s) $D$, initial prompts $\{p_a\}$ for agents $a \in \mathcal{A}$, max iterations $T$
\For{$t = 1$ to $T$}
    \State \textbf{Forward pass:}
    \State \quad Execute RAG pipeline with current prompts $\{p_a\}$:
    \State \qquad -- generate intermediate artifacts (e.g., retrieved chunks or extracted entity--relation graphs)
    \State \qquad -- assemble supporting evidence (e.g., via chunk aggregation or graph community summaries)
    \State \qquad -- produce answer $y_t$
    \State \textbf{Evaluation:}
    \State \quad Evaluator reviews $(y_t, \text{supporting evidence})$ and returns feedback $f_t$ and decision $d_t$
    \If{$d_t=\text{SATISFACTORY}$}
        \State \Return $y_t$
    \EndIf
    \State \textbf{Prompt adaptation:}
    \For{each adaptive agent $a \in \mathcal{A}_{\text{adapt}}$}
        \State \quad Update prompt $p_a \leftarrow \text{Optimize}(p_a, f_t)$
    \EndFor
\EndFor
\State \Return final answer $y_T$
\end{algorithmic}
\end{algorithm}

\subsection{Vector RAG Technical Details}

For flat retrieval, we implement a document-grounded Vector RAG pipeline in which the input document is segmented into overlapping text chunks. A Retrieval Agent follows an IRCoT-style strategy \citep{trivedi-etal-2023-interleaving}, iteratively generating sub-queries based on the context retrieved so far. Each sub-query triggers a hybrid retrieval step that combines dense similarity search and lexical matching to select relevant chunks.

Within \textsc{GradRAG}, the Retrieval Agent is adaptive: its prompt may be updated across refinement cycles based on evaluator feedback. For instance, critiques indicating missing explanations or temporal relationships can lead the agent to generate more targeted sub-queries in subsequent iterations. Retrieved chunks are aggregated to form the context passed to the Answer Generation Agent, as specified in the forward pass of Algorithm~\ref{alg:gradrag}.

\subsection{GraphRAG Technical Details}

Our graph-based RAG pipeline builds on prior graph-oriented approaches \citep{edge2025localglobalgraphrag,pedinotti-santus-2026-structsurvey}. The document is segmented into larger text spans, from which a Graph Extraction Agent identifies entities and relations. Extracted entities are merged into a consolidated graph, and community detection is performed using the Leiden algorithm to identify groups of related entities. For each community, the system generates a natural-language summary and a set of candidate answers, each annotated with a helpfulness score indicating relevance to the query. Only answers with positive scores are retained and concatenated to form the evidence context supplied to the Answer Generation Agent.

As in the vector-based setting, both the Graph Extraction Agent and the Answer Generation Agent are adaptive. Evaluator feedback highlighting missing entities, relations, or explanatory structure is used to refine the extraction prompt, guiding subsequent graph enrichment toward information that is most useful for answering the query. Over successive refinement cycles, this feedback-driven adaptation effectively induces a task-oriented ontology, shaping which types of entities and relations are prioritized during graph construction \citep{pedinotti-santus-2026-structsurvey,pedinotti-etal-2026-metagraph}. The resulting enriched graph is then used to assemble evidence for the final answer in the next forward pass (Algorithm~\ref{alg:gradrag}, lines~2--5).

\paragraph{Implementation.}
All pipelines are implemented using the Microsoft AutoGen framework \citep{wu2023autogenenablingnextgenllm}. The prompts are provided in Appendices~\ref{apx:prompts} and~\ref{apx:prompts_eval} to support reproducibility.

\section{Experimental Setup}
\label{sec:exp}

\paragraph{Research Questions.}
We study whether coordinating prompt adaptation across multiple agents improves RAG pipelines. Specifically, we ask:
(1) Does \textbf{cross-component prompt adaptation} — where evaluator feedback updates upstream agents such as retrievers or graph extractors — improve performance compared to \textbf{one-step refinement} that updates only the final answer generator?
(2) Are these improvements consistent across different RAG architectures, including flat (chunk-based) and graph-based retrieval pipelines?

\paragraph{\textsc{GradRAG} Systems and Baselines.}
To answer these questions, we evaluate \textsc{GradRAG} under two retrieval paradigms: \textbf{flat (chunk-based) retrieval} and \textbf{graph-based retrieval}. For each paradigm, we compare a \textbf{one-step refinement baseline}, in which feedback updates only the answer generator, against a \textbf{full \textsc{GradRAG}} variant that propagates feedback to multiple agents. The systems are:

\begin{itemize}
    \item \textbf{IRCoT \citep{trivedi-etal-2023-interleaving} + One-Step Refinement:}
    A vector-based RAG pipeline in which only the Answer Generation Agent is adaptive. The Retrieval Agent uses a fixed prompt, so improvements arise solely from iterative answer rewriting.

    \item \textbf{Full \textsc{GradRAG} Vector:}
    The same pipeline, but evaluator feedback updates both the Answer Generation Agent and the Retrieval Agent, enabling refinement of sub-query generation and evidence selection.

    \item \textbf{GraphRAG \citep{edge2025localglobalgraphrag} + One-Step Refinement:}
    A graph-based RAG pipeline where only the Answer Generation Agent is adaptive. The Graph Extraction Agent uses a fixed prompt, so improvements arise only from refining the final answer.

    \item \textbf{Full \textsc{GradRAG} Graph:}
    The same pipeline, but evaluator feedback updates both the Answer Generation Agent and the Graph Extraction Agent, enabling adaptive graph enrichment.
\end{itemize}

This setup forms a controlled ablation: within each retrieval paradigm, the baseline and GradRAG variants share the same architecture, model, and retrieval process, differing only in whether evaluator feedback updates upstream agents. This isolates the effect of cross-component prompt adaptation.

\paragraph{Task.}
We evaluate all systems on the \textbf{Query-Focused Summarization} (QFS) task. In QFS, the model receives a document (or document set) $d$ and a query $q$, and must produce a summary that answers the query using information drawn from the document. The task lies at the intersection of abstractive summarization and question answering, requiring both global document understanding and targeted evidence selection.

\paragraph{Datasets.}
We conduct experiments on two standard QFS datasets: \textbf{\textsc{SQuALITY}} \citep{wang-etal-2022-squality} and \textbf{\textsc{QMSum}} \citep{zhong-etal-2021-qmsum}. These datasets represent complementary domains: narrative texts (\textsc{SQuALITY}) and long meeting transcripts (\textsc{QMSum}). Both datasets are also included in the ZeroSCROLLS \citep{shaham-etal-2023-zeroscrolls} and LongBench \citep{bai-etal-2024-longbench} benchmarks. Although these documents fit within the context windows of modern LLMs, we use these datasets as controlled benchmarks for studying the behavior of RAG pipelines and evaluating coordination between their components. Table~\ref{tab:dataset_informations} summarizes their main characteristics.

\begin{table}[t]
\centering
\resizebox{0.48\textwidth}{!}{%
\begin{tabular}{lrrrr}
\toprule
\textbf{Dataset} & \textbf{Docs} & \textbf{Questions} & \textbf{Avg Doc Length} & \textbf{Avg Resp Length} \\
\midrule
\textsc{SQuALITY} & 52 & 260 & 4,995 words & 241 words \\
\addlinespace
\textsc{QMSum} & 35 & 281 & 10,668 words & 65 words \\
\bottomrule
\end{tabular}%
}
\caption{Statistics about the test sets.}
\label{tab:dataset_informations}
\end{table}

\paragraph{Experimental Protocol.}
\textsc{GradRAG} operates in a \textbf{test-time refinement} setting. For each test instance (query–document pair), the system executes an iterative loop consisting of a forward pass followed by evaluation and, if necessary, prompt adaptation. During refinement, an \emph{Evaluator Agent} analyzes the generated answer together with its supporting evidence and produces structured feedback identifying missing information or reasoning gaps. Importantly, the Evaluator does \emph{not} have access to the reference answer, preventing data leakage during optimization.

If the Evaluator determines that the answer is unsatisfactory, its critique is used to update the prompts of the adaptive agents, and another refinement cycle is executed. Otherwise, the process terminates through an \textbf{early-stopping} decision. Each test instance is processed independently and optimized prompts are reset after each example, ensuring that no information is carried across test instances.

This protocol enables a controlled comparison between one-step refinement and cross-component adaptation by isolating the effect of prompt updates within a single query. While this per-instance refinement increases computational cost, it avoids cross-example contamination. In deployment scenarios, efficiency could be improved by batching multiple instances before updating prompts, which we leave for future work.

\paragraph{Evaluation Metrics.}
Traditional evaluation metrics for query-focused summarization, such as ROUGE and BERTScore, correlate poorly with human judgments because they primarily measure lexical or embedding overlap rather than higher-level properties such as coherence, factuality, and reasoning \citep{kryscinski-etal-2019-neural, liu-etal-2023-g, nguyen2024comparativestudyqualityevaluation}. 

During refinement, \textsc{GradRAG} uses an internal \emph{Evaluator Agent} that analyzes generated answers together with their supporting evidence and produces critiques guiding prompt updates. This evaluator operates without access to gold reference answers to prevent data leakage. For final system comparison, we instead adopt an \textbf{LLM-as-a-Judge} evaluation paradigm in which an external LLM compares system outputs using a structured rubric. Given two candidate answers for the same input, the judge selects the better response based on overall answer quality and use of supporting evidence. We report \textbf{win rates}, defined as the proportion of pairwise comparisons in which one system is preferred over the other \citep{liu2025aligninghumanjudgementrole}.

To improve evaluation robustness, we follow recommendations from recent work on LLM-based evaluation \citep{baumann2025largelanguagemodelhacking}. Specifically, we report \textbf{position consistency} \citep{shi2025judgingjudgessystematicstudy}, which measures how often the judge's preference remains unchanged when the order of candidate answers is reversed. We also apply a two-sided \textbf{binomial sign test} to assess statistical significance.

We also conduct \textbf{human validation} on a subset of examples. Annotators are shown the same pairwise outputs and asked to select the better response using the same rubric. Agreement between LLM and human preferences is measured using accuracy and Cohen’s $\kappa$ computed on majority-vote labels.

Finally, to analyze refinement dynamics (Section~\ref{sec:res}), we track several per-iteration diagnostics: \textbf{average response length}, \textbf{lexical density} (content-word ratio), \textbf{ROUGE-1} between the system response and the gold reference answer, and \textbf{topic entropy}, defined as the Shannon entropy of topic distributions obtained by fitting LDA to the set of system responses at each iteration.

\paragraph{Implementation Details and Hyperparameters.}
\label{para:hyperparameters}

For \textbf{flat (chunk-based) retrieval}, documents are segmented into overlapping text chunks using fixed character lengths (default 400 characters with a 40-character overlap). Retrieval follows an IRCoT-style strategy \citep{trivedi-etal-2023-interleaving}, in which the Retrieval Agent iteratively generates sub-queries based on the context retrieved so far. Each sub-query triggers a retrieval step, where the Retrieval Agent selects the retrieval strategy to apply: dense similarity search via FAISS \citep{douze2025faisslibrary} or lexical matching via BM25 \citep{robertson1995okapi}. We perform three retrieval iterations and select the top retrieved chunks at each step, aggregating them to form the final context passed to the Answer Generation Agent. The choice of retrieval strategy is controlled by the agent’s prompt and may change across refinement iterations as a result of evaluator feedback.

For \textbf{graph-based retrieval}, documents are segmented into larger overlapping spans (default 1000 characters with a 200-character overlap) to support more reliable entity and relation extraction. A Graph Extraction Agent identifies entities and relations from these spans, which are aggregated into a graph structure that is incrementally enriched across refinement iterations. The resulting graph is then used to assemble evidence for answer generation. Across all systems, the maximum number of refinement iterations is capped at three.

\paragraph{Models.}
All generation agents (Retriever, Graph Extractor, and Answer Generator) use \textbf{Gemini-2.5-Flash} \citep{comanici2025gemini25pushingfrontier} (June 2025 version) via the Google Generative Language API.\footnote{\url{https://ai.google.dev/gemini-api/docs/models}} 
Repetitive subtasks—such as sub-query generation in flat retrieval, entity–relation extraction in graph retrieval, and community-level summarization and candidate-answer generation—use the more cost-efficient \textbf{Gemini-2.5-Flash-Lite} (July 2025 version) through the same API. For vector indexing, we use the \textbf{text-embedding-3-small} embedding model.\footnote{\url{https://platform.openai.com/docs/guides/embeddings}} 
For evaluation, we use the open \textit{DeepSeek-V3.1} model \citep{deepseekai2025deepseekv3technicalreport} as an external LLM judge. The prompts used by the system and the evaluator are provided in Appendices~\ref{apx:prompts} and~\ref{apx:prompts_eval} to support reproducibility.

\paragraph{Computational Cost.}
We analyze the additional overhead introduced by cross-component prompt adaptation using the GraphRAG pipeline on a sample of 13 query–document pairs. A full refinement cycle requires on average 35s per query compared to 30.5s for the one-step baseline, corresponding to roughly a 10\% increase. Most runtime (about 70\%) is spent in the forward pipeline—graph extraction and answer generation—which is identical across systems. Token usage increases from 6,040 tokens per query in the baseline to 8,140 tokens in \textsc{GradRAG}, primarily due to critique processing during prompt updates.

\begin{table}[htbp]
\centering
\small
\renewcommand{\arraystretch}{1.35}
\begin{tabular}{p{0.48\columnwidth}cc}
\toprule
\textbf{Retrieval Paradigm} & \textbf{\textsc{SQuALITY}} & \textbf{\textsc{QMSum}} \\
\midrule
\textbf{Flat (Vector RAG)} & & \\
Full \textsc{GradRAG} win rate (\%) & 56.5 & 56.0 \\
Position consistency (\%) & 79.3 & 55.1 \\
$p$-value & 0.028 & 0.090 \\
\midrule
\textbf{Graph-based RAG} & & \\
Full \textsc{GradRAG} win rate (\%) & 57.5 & 55.8 \\
Position consistency (\%) & 82.5 & 87.9 \\
$p$-value & 0.005 & 0.020 \\
\bottomrule
\end{tabular}
\caption{Pairwise evaluation results. Values report the percentage \textbf{win rate} of \textsc{GradRAG} against the corresponding one-step baseline.}
\label{tab:main_res}
\end{table}

\begin{table}[t]
\centering
\small
\renewcommand{\arraystretch}{1.25}
\begin{tabular}{p{0.48\columnwidth}cc}
\toprule
\textbf{Metric} & \textbf{\textsc{SQuALITY}} & \textbf{\textsc{QMSum}} \\
\midrule
Full \textsc{GradRAG} win rate (\%) & 58.5 & 61.5 \\
Position consistency (\%) & 79.3 & 59.7 \\
$p$-value & 0.0038 & 0.0007 \\
\bottomrule
\end{tabular}
\caption{Effect of an additional refinement iteration for flat retrieval.}
\label{tab:additional_iteration_QMSum}
\end{table}

\begin{table}[htbp]
\centering
\small
\renewcommand{\arraystretch}{1.3}
\begin{tabular}{p{0.48\columnwidth}cc}
\toprule
\textbf{Method} & \textbf{\textsc{SQuALITY}} & \textbf{\textsc{QMSum}} \\
\midrule
\multicolumn{3}{c}{\textit{Flat Retrieval}} \\
IRCoT + One-step refinement & 1.82 & 1.60 \\
Full \textsc{GradRAG} Vector & 1.84 & 1.63 \\
\midrule
\multicolumn{3}{c}{\textit{Graph Retrieval}} \\
GraphRAG + One-step refinement & 1.83 & 1.63 \\
Full \textsc{GradRAG} Graph & 1.72 & 1.68 \\
\bottomrule
\end{tabular}
\caption{Average number of refinement iterations before early stopping.}
\label{tab:mean_iterations_per_system}
\end{table}

\begin{figure*}[!b]
%\begin{figure*}[t] 
\centering \begin{subfigure}{0.24\linewidth} \centering \includegraphics[width=\linewidth]{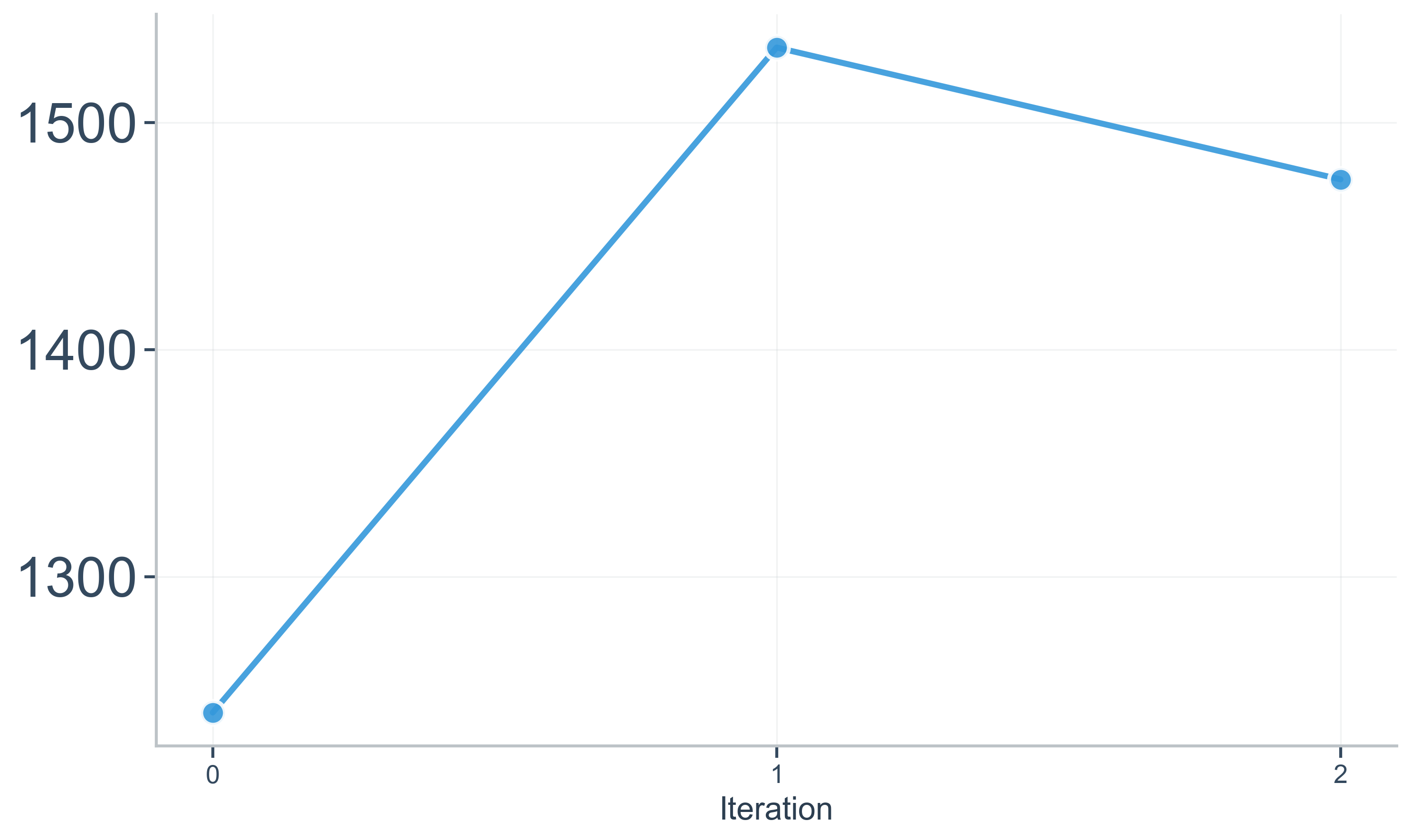} \caption{Length} \end{subfigure} \begin{subfigure}{0.24\linewidth} \centering \includegraphics[width=\linewidth]{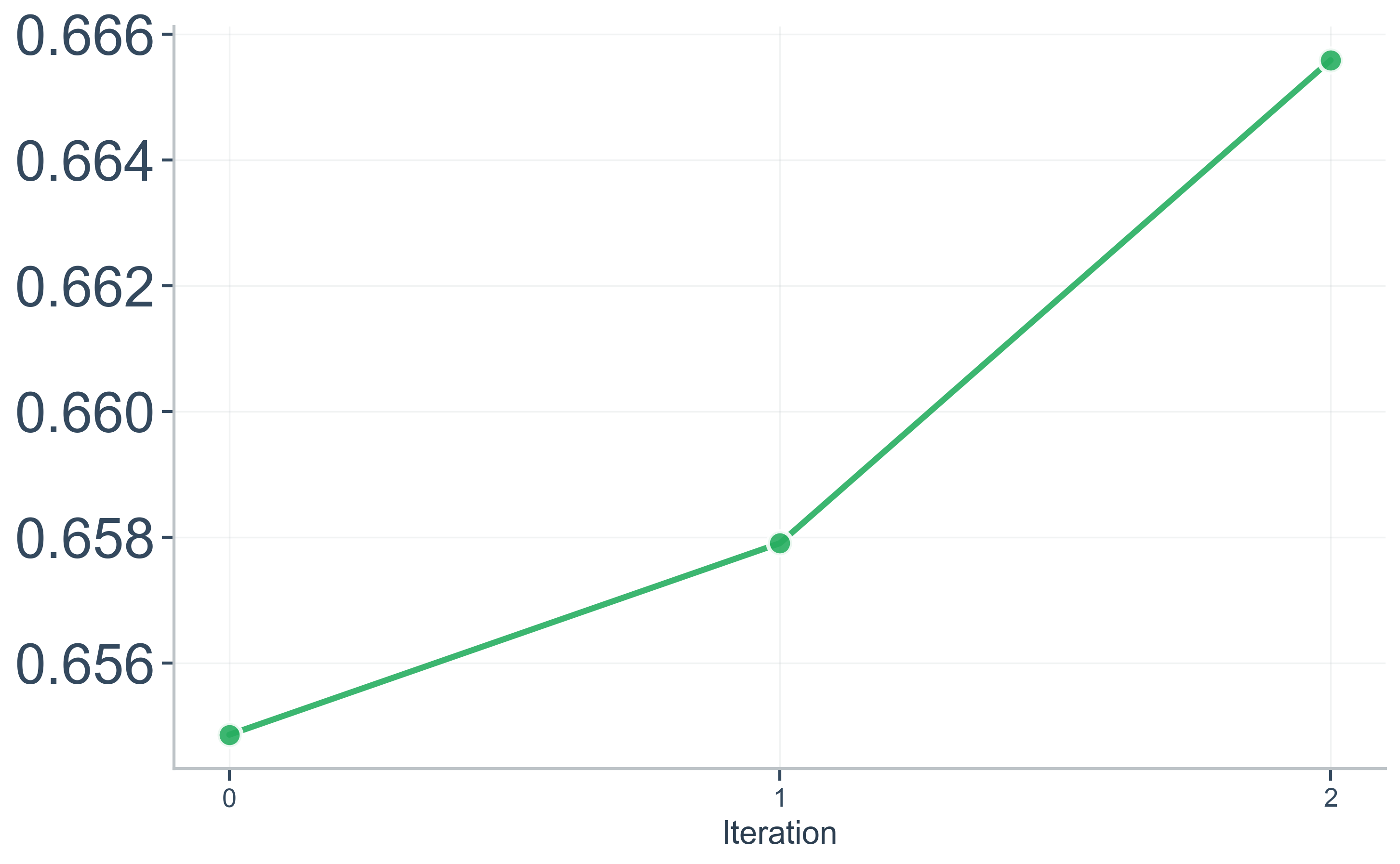} \caption{Lexical density} \end{subfigure} \begin{subfigure}{0.24\linewidth} \centering \includegraphics[width=\linewidth]{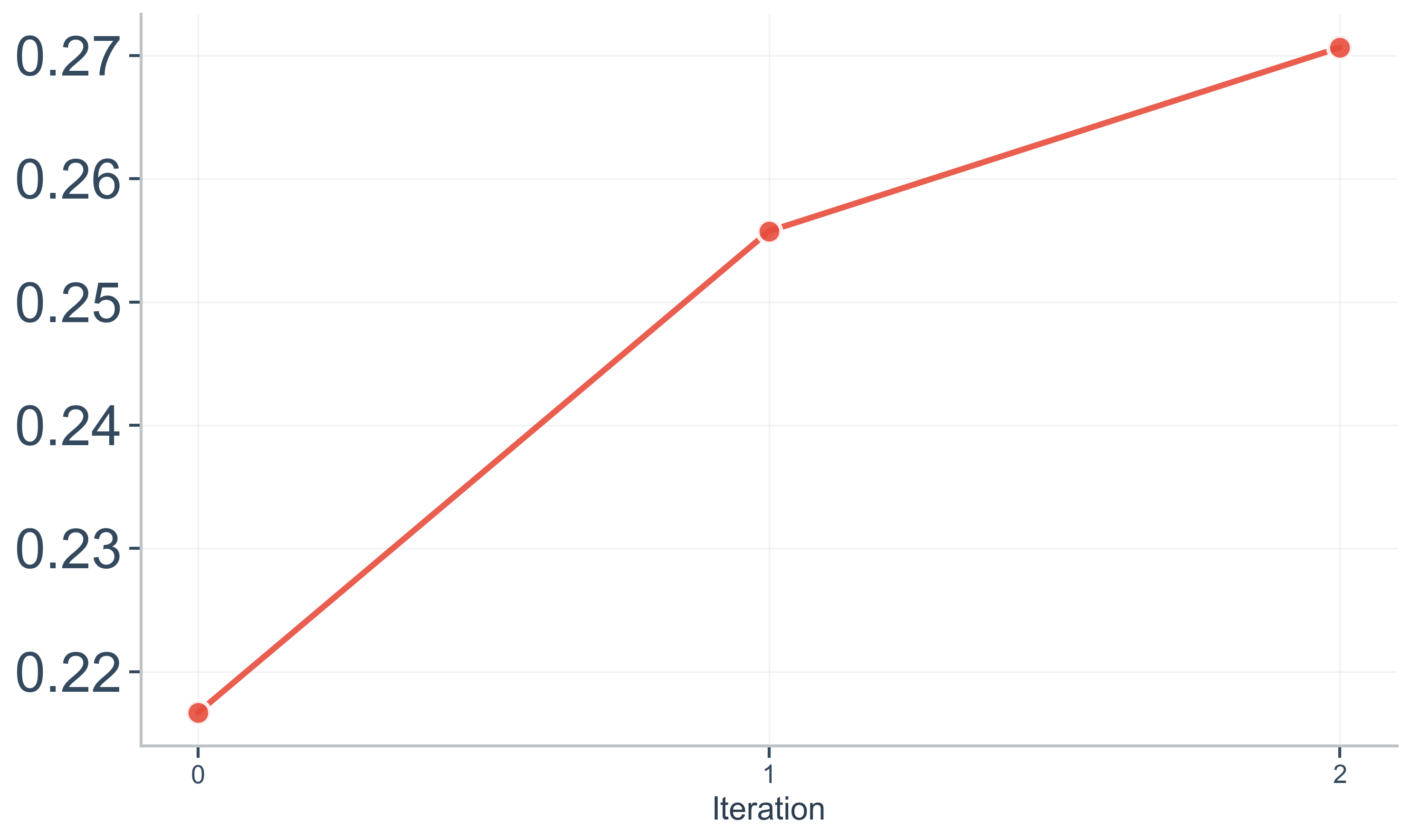} \caption{ROUGE-1} \end{subfigure} \begin{subfigure}{0.24\linewidth} \centering \includegraphics[width=\linewidth]{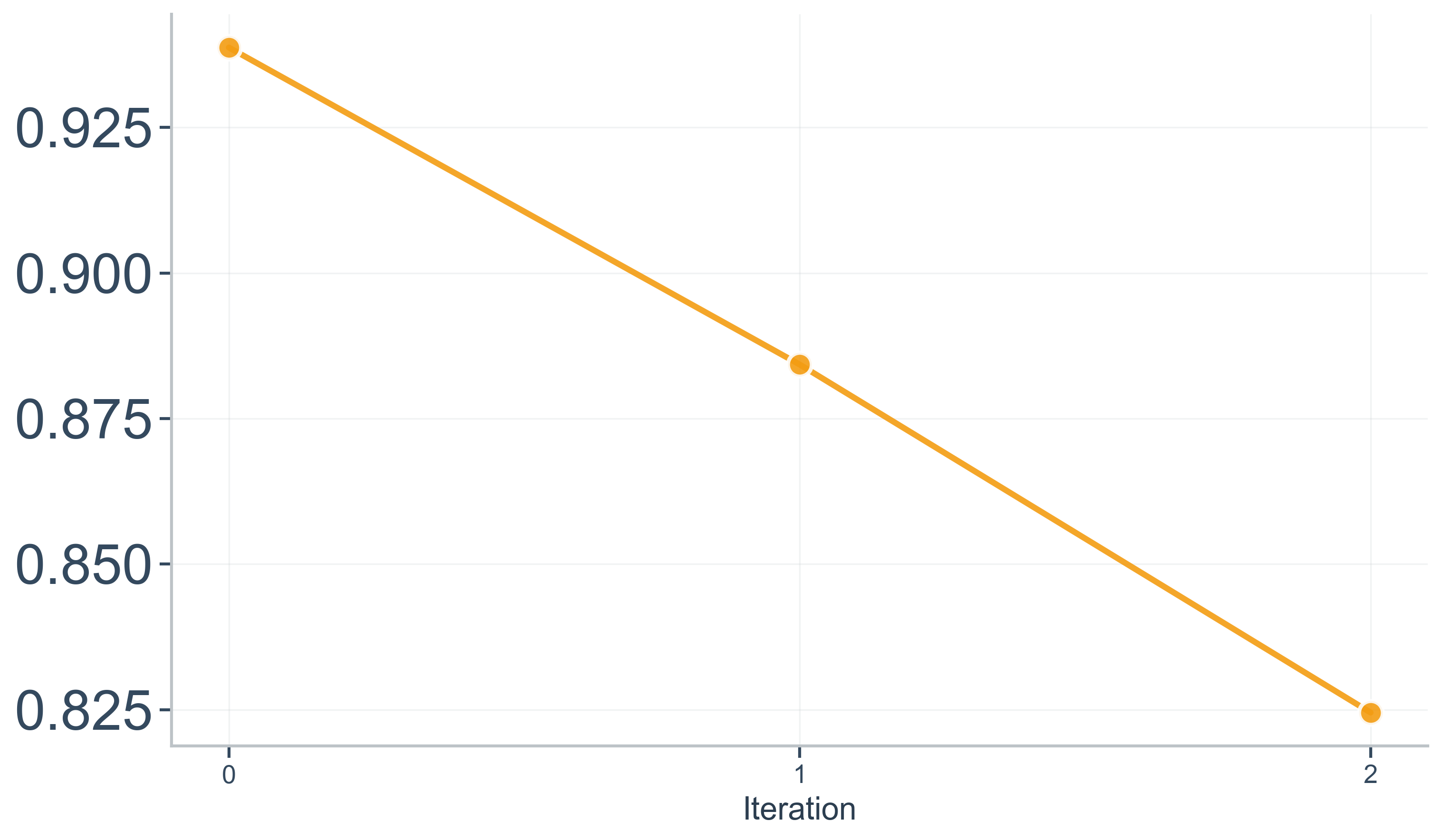} \caption{Topic entropy} \end{subfigure} \caption{Evolution of system responses across iterations.} \label{fig:evolution_responses} \end{figure*}

\begin{table}[htbp]
\centering
\small
\renewcommand{\arraystretch}{1.3}
\begin{tabular}{p{0.48\columnwidth}cc}
\toprule
\textbf{Retrieval Paradigm} & \textbf{\textsc{SQuALITY}} & \textbf{\textsc{QMSum}} \\
\midrule
\multicolumn{3}{c}{\textit{Iteration 2}} \\
Flat (Vector) GradRAG win rate (\%) & 54.0 & 54.6 \\
Graph-based GradRAG win rate (\%) & 53.6 & 55.6 \\
\midrule
\multicolumn{3}{c}{\textit{Iteration 3}} \\
Flat (Vector) GradRAG win rate (\%) & 56.5 & 56.0 \\
Graph-based GradRAG win rate (\%) & 57.5 & 55.8 \\
\bottomrule
\end{tabular}
\caption{Pairwise preference rates for \textsc{GradRAG} over one-step baselines when the number of refinement iterations is fixed to a given value (without early stopping).}
\label{tab:early_results}
\end{table}

\section{Results}
\label{sec:res}

%We first report the alignment between LLM-based preference judgments and human annotations, as defined in the evaluation protocol (Section~\ref{sec:exp}). We then present the main experimental results comparing one-step baselines and full \textsc{GradRAG} variants, followed by an analysis of refinement dynamics and response evolution across iterations.

\paragraph{Human Validation.}
Human evaluation results show substantial agreement between LLM-based judgments and human preferences. Across the annotated subset, the LLM achieves a $75.86\%$ agreement rate with human evaluators, a Cohen’s~$\kappa$ of $0.59$, and an F1 score of $0.8$. Human preferences exhibit moderate variability, with an average annotation entropy of $0.642$, indicating partial disagreement on more challenging examples, particularly those involving longer responses. %Overall, these results suggest that the LLM judge’s preferences are broadly aligned with human judgments for this task.

\paragraph{Main Results.}
Table~\ref{tab:main_res} reports pairwise comparison results between one-step baselines and full \textsc{GradRAG} variants. Comparisons are performed within each retrieval paradigm (flat vs.\ graph-based), isolating the effect of cross-component prompt adaptation from differences in retrieval architecture.

Across both datasets and retrieval paradigms, full \textsc{GradRAG} variants consistently outperform their one-step counterparts. Win rates range between $55\%$ and $57\%$, corresponding to a \textbf{net preference margin} of approximately $12$–$15$ percentage points when compared with the corresponding baseline preferences (e.g., $56\%$ vs.\ $44\%$). These results indicate that allowing evaluator feedback to update multiple upstream agents yields consistent improvements over refining only the final answer generator.

Evaluation robustness diagnostics are reported alongside preferences (Section~\ref{sec:exp}). Three of the four comparisons yield statistically significant differences at the $p < 0.05$ level. The remaining case—flat retrieval on \textsc{QMSum}—shows marginal significance, reflecting greater variability in evaluator preferences for this setting.

To further investigate this case, we run an additional refinement iteration for flat retrieval on \textsc{QMSum}. As shown in Table~\ref{tab:additional_iteration_QMSum}, the additional iteration increases the performance gap and yields statistically significant results, suggesting that the earlier marginal outcome was likely due to insufficient refinement rather than instability in the evaluation procedure.

\paragraph{Effect of Refinement Budget.}
Table~\ref{tab:mean_iterations_per_system} reports the average number of refinement iterations before early stopping. Across systems, refinement typically terminates after fewer than two iterations, indicating that satisfactory outputs are often reached well before the maximum iteration budget.

Table~\ref{tab:early_results} reports pairwise preference rates as a function of a fixed number of refinement iterations, evaluated without early stopping. For flat (chunk-based) retrieval, performance generally improves as additional refinement iterations are permitted, with win rates increasing from iteration~2 to iteration~3 on both datasets. Graph-based \textsc{GradRAG} shows a similar pattern, achieving preference rates above $50\%$ at both iteration limits across datasets. While gains are not strictly monotonic in all cases—most notably on \textsc{QMSum}, where the win rate slightly decreases from iteration~2 to iteration~3—the graph-based variant remains consistently preferred over the one-step baseline.

Overall, these results suggest that both flat and graph-based pipelines benefit from additional refinement under fixed budgets, while early stopping helps balance refinement gains against potential over-specialization in structured retrieval settings.

\paragraph{Response Evolution Across Iterations.}
Figure~\ref{fig:evolution_responses} illustrates how system outputs evolve across refinement iterations. Response length increases sharply after the first refinement and then stabilizes, while lexical density increases steadily across iterations. ROUGE-1 scores improve monotonically, and topic entropy decreases over time, indicating progressively more focused responses.

Taken together, these trends suggest that refinement primarily improves responses by increasing informational density and topical focus rather than by simply adding more content. This pattern is consistent across retrieval paradigms and supports the interpretation that cross-component prompt adaptation produces more targeted and coherent outputs.

\section{Conclusion}
\label{sec:conc}

We introduced \textbf{\textsc{GradRAG}}, a framework for coordinating multiple agents in Retrieval-Augmented Generation (RAG) pipelines through cross-component prompt adaptation. By modeling RAG systems as computational graphs and propagating evaluation feedback across agents, \textsc{GradRAG} enables coordinated refinement of retrieval, evidence construction, and answer generation at test time.

Across both vector- and graph-based retrieval settings, \textsc{GradRAG} achieves a \textbf{12--15 percentage point net preference margin} in pairwise comparisons. These results indicate that allowing evaluation feedback to influence upstream components—rather than refining only the final answer—consistently improves system outputs. More broadly, our findings highlight the importance of coordinating adaptation across heterogeneous components in modular RAG pipelines.

\section{Limitations}
\label{sec:lim}
While \textsc{GradRAG} introduces a general framework for agentic optimization in RAG systems, several aspects offer opportunities for further refinement:

\begin{itemize}[noitemsep, topsep=0pt]
    \item \textbf{Computational cost.} The approach is computationally intensive, as both the forward execution and refinement phases rely on LLM generation. However, this cost scales predictably with the number of adaptive agents and remains manageable for moderate configurations. Moreover, efficiency gains can be achieved through parallelization.
    
    \item \textbf{Evaluation vs.\ deployment gap.} Our evaluation protocol — where each refinement cycle is triggered after a single forward execution — was designed to ensure data isolation and interpretability. While this setup is less reflective of deployment conditions, it provides a clean benchmark for studying agentic feedback dynamics. In contrast, recent approaches to automatic prompt optimization, such as DSPy \citep{khattab2023dspycompilingdeclarativelanguage}, derive optimized prompts from multiple examples. Future research may extend \textsc{GradRAG} to batch- or continual-feedback regimes to better approximate real-world usage.
    
    \item \textbf{Evaluator configuration.} The Evaluator Agent operates using general summarization principles (e.g., coherence, fluency) without external reward signals. This design isolates the intrinsic capability of LLMs to self-evaluate while avoiding data leakage. Future implementations could integrate additional supervision signals, such as reward models or heuristic metrics, to strengthen evaluation fidelity.
\end{itemize}

%\section*{Acknowledgments}
%We thank David Rosenberg for his detailed and insightful feedback, which substantially improved this paper, and Ioana Baldini, Kay Nandwani, and Pawel Polak for their thoughtful reviews and helpful suggestions.

%\section{Conclusion}

%In this work, we introduced \textsc{GradRAG}, a general framework for building self-improving RAG systems through global feedback propagation, coordinated multi-agent optimization, and modular computational design. Across both vector-based and graph-based retrieval settings, our experiments show that enabling feedback to flow backward through the entire pipeline—rather than remaining confined to local components—consistently improves system output. By ensuring that agents adapt in coordination rather than in isolation, \textsc{GradRAG} supports collective refinement, where upstream and downstream components evolve toward shared objectives. Finally, the modular abstraction of the pipeline as a computational graph allows the same optimization principles to be applied seamlessly across heterogeneous architectures, highlighting the generality of the approach. Together, these results demonstrate that global propagation, coordinated adaptation, and modular composition are not only theoretically compelling but also practically effective design principles for next-generation agentic RAG systems.

\bibliography{custom}

\appendix

\section{Prompts for Evaluator Agents}
\label{apx:prompts}
\subsection{Flat Retrieval}

\subsubsection*{Response evaluator agent}

\begin{lstlisting}[style=promptstyle]
You are an expert judge evaluating both the retrieved context and the generated answer for completeness, coherence, and overall quality.

Your task:
1. Context Evaluation:
    - First, imagine an ideal, fully satisfying answer to the question - one that captures all key facts, causal and temporal connections, character motivations, outcomes, and thematic meaning.
   - Then compare this ideal answer with the retrieved context. Identify what information, links, or nuances are missing, underdeveloped, ambiguous, or irrelevant.
   - Note any missing elements that would prevent a reader from forming a complete understanding or that weaken the logical or emotional flow of the narrative.

2. Answer Evaluation:
   - Assess whether the answer uses the available context effectively.
   - Evaluate clarity, coherence, factual accuracy, completeness, and narrative flow.
   - Check if it connects events and entities logically, avoids contradictions, and reads fluently in tone and length.

Decision:
- SATISFACTORY: The context includes all major information and connections needed for an ideal answer, and the answer expresses them clearly and coherently.
- NEEDS_REFINEMENT: Any important fact, relation, or narrative link is missing, unclear, or the answer lacks fluency, structure, or accuracy.

Critique must include:
1. What an ideal answer would contain that is not fully supported by the context.
2. Which parts of the context are irrelevant or insufficient.
3. How the answer could improve in factual precision, logical or causal connections, completeness, and stylistic quality.

Output format:
DECISION: [SATISFACTORY or NEEDS_REFINEMENT]
CRITIQUE: [detailed explanation]

The question was the following:
{original_query}

The answer of the system was:
{generated_answer}

When generating the answer, the system had access to the following context:
{retrieved_context}
\end{lstlisting}

\subsubsection*{Answer Generation Agent critique prompt}

\begin{lstlisting}[style=promptstyle]
You are evaluating the prompt used for answer generation in a RAG system.
Your goal is to evaluate the prompt and propose actionable improvements based on feedback obtained from the system's output.
The feedback contains strategies to improve both the retrieved content and the style of the answer; focus only on the style.

Current answer generation prompt:

{current_prompt}

Based on this information, determine if there is a problem with the answer generation prompt that needs to be fixed.

Provide a detailed critique to the prompt reporting what you learned in the previous critique. The critique must be actionable: it must include concrete actions that would likely improve the output.
If false, leave the critique empty.
\end{lstlisting}

\subsubsection*{Answer generation prompt optimizer prompt}

\begin{lstlisting}[style=promptstyle]
You are optimizing a prompt for answer generation in a RAG system. The prompt must be general, but it also has to be as adherent as possible to the following critique.

The current prompt is:
{current_prompt}

The current critique of the answer generation process is:
{answer_generation_critique}

Based on this critique, generate a new prompt that will be used to instruct the LLM how to better generate answers from retrieved vector information. The prompt should incorporate the feedback to improve answer quality, relevance, and coherence.

Provide only the optimized prompt without additional commentary.
"""
\end{lstlisting}

\subsubsection*{Retrieval agent critique prompt}

\begin{lstlisting}[style=promptstyle]
You are evaluating the queries made by an iterative content retriever in a RAG system.
Your goal is to provide an accurate evaluation of the queries based on the following feedback.

The original question of the user was:
{question}

QUERIES MADE:
{queries_formatted}

RESPONSE EVALUATOR FEEDBACK:
{response_evaluator_critique}

The previous critique for the  answer generator was:
{answer_generator_critique}

Based on the response evaluator's feedback, provide:

1. TYPES OF QUERIES TO FOCUS ON:
   - What types of queries would address the issues mentioned in the feedback?
   - What aspects should future queries target to improve the answer?
   - Remember that each query must target a specific part of the text. A good query targets specific content that can be found in a single chunk.

Your task is not to rate the queries, but to identify what types of new queries would best address the issues raised in the feedback.
Relate your suggestions directly to the user original question and the retrieved content, specifying what missing details or narrative connections new queries should aim to uncover.
Provide a clear, actionable analysis.
\end{lstlisting}

\subsubsection*{Retrieval agent prompt optimizer}

\begin{lstlisting}[style=promptstyle]
You are optimizing a prompt for a query agent in a vector RAG system.

The current prompt is:
{current_prompt}

IMPORTANT CLARIFICATION:
- The planner you are optimizing asks the model to generate a sub-query to answer the question, based on the context retrieved so far and the previous sub-queries.
- The planner does NOT control the retrieval/embedding mechanism itself
- Your instructions should focus on which aspects a sub-query must target

You will receive:
1. The original question to answer
2. The sub-queries that were made
3. A critique of the retrieval plan identifying what worked and what needs to be improved.

{original_question}

{previous_queries}

{retrieval_plan_critique}

Based on this information, generate an optimized prompt that instructs the query agent to create a good sub-query.


Provide only the optimized prompt as instructions for the query planner, without additional commentary.

Generate a full optimized prompt (not partial additions) that will replace the current one used by the query agent.

\end{lstlisting}

\subsection{Graph Retrieval}

\subsubsection*{Response evaluator agent}

\begin{lstlisting}[style=promptstyle]
You are an expert judge evaluating both the retrieved context and the generated answer for completeness, coherence, and overall quality.

Your task:
1. Context Evaluation:
   - First, imagine an ideal, fully satisfying answer to the question - one that captures all key facts, causal and temporal connections, character motivations, outcomes, and thematic meaning.
   - Then compare this ideal answer with the retrieved context. Identify what information, links, or nuances are missing, underdeveloped, ambiguous, or irrelevant.
   - Note any missing elements that would prevent a reader from forming a complete understanding or that weaken the logical or emotional flow of the narrative.

2. Answer Evaluation:
   - Assess whether the answer uses the available context effectively.
   - Evaluate clarity, coherence, factual accuracy, completeness, and narrative flow.
   - Check if it connects events and entities logically, avoids contradictions, and reads fluently in tone and length.

Decision:
- SATISFACTORY: The context includes all major information and connections needed for an ideal answer, and the answer expresses them clearly and coherently.
- NEEDS_REFINEMENT: Any important fact, relation, or narrative link is missing, unclear, or the answer lacks fluency, structure, or accuracy.

Critique must include:
1. What an ideal answer would contain that is not fully supported by the context.
2. Which parts of the context are irrelevant or insufficient.
3. How the answer could improve in factual precision, logical or causal connections, completeness, and stylistic quality.

Output format:
DECISION: [SATISFACTORY or NEEDS_REFINEMENT]
CRITIQUE: [detailed explanation]

The question was the following:
{original_query}

The answer of the system was:
{generated_answer}

When generating the answer, the system had access to the following context:
{retrieved_context}
\end{lstlisting}

\subsubsection*{Answer generation critique prompt}

\begin{lstlisting}[style=promptstyle]
You are evaluating the prompt used for answer generation in a GraphRAG system.
Your goal is to evaluate the prompt and propose actionable improvements based on feedback obtained from the system's output.
The feedback contains strategies to improve both the retrieved content and the style of the answer; FOCUS ONLY ON THE STYLE.

Current answer generation prompt:
{current_prompt}

Feedback from the response evaluation:
{response_evaluator_output}

Based on this information, determine if there is a problem with the answer generation prompt that needs to be fixed.

First, provide your reasoning explaining why there is or isn't a problem.
Then, provide a critique focusing on the specific issue.
\end{lstlisting}

\subsubsection*{Answer generation prompt optimizer}

\begin{lstlisting}[style=promptstyle]
You are optimizing a prompt for answer generation in a GraphRAG system.

The current critique of the answer generation process is:
{answer_generation_critique}

Based on this critique, generate a new prompt that will be used to instruct the LLM how to better generate answers from retrieved graph information. The prompt should incorporate the feedback to improve answer quality, relevance, and coherence.

Provide only the optimized prompt without additional commentary.
\end{lstlisting}

\subsubsection*{Graph Extraction agent critique prompt}

\begin{lstlisting}[style=promptstyle]
You are evaluating the prompt given to an LLM to enrich a given graph with the information extracted from the text.
The prompt tells the system how to extract entities and relationships.
You will be provided with feedback explaining which information is missing in the graph.
You have to think: which entity/relationship types would make this information available in the graph?
Clearly specify the entities and relationship types in your answer.
Include only a few entities and relationship types (not more than 6-7).
Based on this, you have to identify entity and relationship types that should be included in the graph and are not specified in the current prompt.
Focus only on the most crucial entity/relationship types to meet the evaluation requirement. Specify only a few entity/relationship types, the ones that are most important.

For each entity/relationship, you have to include examples (each example is a phrase or a sentence).

The feedback from the system is:
{response_evaluator_output}

The previous critique for the  answer generator was:
{answer_generator_critique}

The current prompt is:
{current_prompt}

Please, determine which entities/relationships the current prompt is missing based on the feedback.

\end{lstlisting}

\subsubsection*{Graph Extraction agent prompt optimizer}

\begin{lstlisting}[style=promptstyle]
Your goal is to generate an instruction for an LLM that enriches a graph using information from the text.
The instruction must have the following format: "Focus on:" + LIST OF ENTITY/RELATIONSHIP TYPES
The instruction tells the model which entity/relationship types to expand the graph with.
Include only A FEW entities and relationship types (NOT MORE THAN 6-7 IN TOTAL).
You have to accompany each entity/relationship with a full explanation and some examples. Each example is a phrase or a sentence.
Specify only the relationships and the entities, don't use formatted examples since they can mislead the model output format.

Your suggestions must be based on this feedback:
{graph_extraction_feedback}

Provide only the instruction without additional commentary.
\end{lstlisting}

\section{LLM-as-a-Judge Evaluation Details}
\label{apx:prompts_eval}
\subsection*{Model}

\begin{itemize}
    \item Model: \texttt{deepseek-ai/DeepSeek-V3.1} with thinking mode enabled
    \item Size: 685B parameters
\end{itemize}

\subsection*{Prompt Template}

\begin{lstlisting}[style=promptstyle]
You are evaluating two answers to a question. Your task is to determine which answer better adheres to the gold-standard reference
answer.

**Question:**
{question}

**Gold Standard Reference Answer:**
{gold_standard}

**Answer A (from {system_a}):**
{answer_a}

**Answer B (from {system_b}):**
{answer_b}

**Instructions:**
1. First, provide your reasoning about which answer is better and why. 

2. After your reasoning, you MUST end your response with EXACTLY one of these options on the last line, with nothing else:
   A
   B

Do not add any punctuation, explanation, or other text on the final line. Just the single letter or word.
\end{lstlisting}

\end{document}